\documentclass[pdflatex,sn-mathphys-num]{sn-jnl}

\usepackage{graphicx}%
\usepackage{multirow}%
\usepackage{amsmath,amssymb,amsfonts}%
\usepackage{amsthm}%
\usepackage{mathrsfs}%
\usepackage[title]{appendix}%
\usepackage{xcolor}%
\definecolor{commentgreen}{RGB}{34, 139, 34} 
\usepackage{textcomp}%
\usepackage{manyfoot}%
\usepackage{booktabs}%
\usepackage{algorithm}%
\usepackage{algorithmicx}%
\usepackage{algpseudocode}%
\usepackage{listings}%

\theoremstyle{thmstyleone}%
\theoremstyle{thmstyletwo}%

\theoremstyle{thmstylethree}%

\begin{document}

\title[Article Title]{The Epochal Sawtooth Phenomenon: Unveiling Training Loss Oscillations in Adam and Other Optimizers}


\author*[1]{\fnm{Qi} \sur{Liu}}\email{liu\_qi@tongji.edu.cn}

\author[2]{\fnm{Wanjing} \sur{Ma}}\email{mawanjing@tongji.edu.cn}

\affil[1,2]{\orgdiv{College of Transportation}, \orgname{Tongji University}, \orgaddress{\street{4800 Cao'an Rd.}, \city{Shanghai}, \postcode{201804}, \state{Shanghai}, \country{P.R.China}}}


\abstract{In this paper, we identify and analyze a recurring training loss pattern, which we term the \textit{Epochal Sawtooth Phenomenon (ESP)}, commonly observed during training with adaptive gradient-based optimizers, particularly Adam optimizer. This pattern is characterized by a sharp drop in loss at the beginning of each epoch, followed by a gradual increase, resulting in a sawtooth-shaped loss curve. Through empirical observations, we demonstrate that while this effect is most pronounced with Adam, it persists, although less severely, with other optimizers such as RMSProp. We empirically analyze the mechanisms underlying ESP, focusing on key factors such as Adam's $\beta$ parameters, batch size, data shuffling, and sample replacement. Our analysis shows that ESP arises from adaptive learning rate adjustments controlled by the second moment estimate. Additionally, we identify the ``immediate re-exposure to samples'' effect during data shuffling, which causes the model to learn or memorize more at the beginning of each epoch. We also find that smaller values of $\beta_2$ exacerbate ESP but can act as a form of regularization. While ESP is not necessarily indicative of overfitting, higher model capacity can amplify the phenomenon. To further support our analysis, we replicate ESP through a high-dimensional quadratic minimization task. We demonstrate that ESP can emerge even in simple optimization scenarios, reinforcing the generality of this pattern. The code for reproducing our experiments is available at \url{https://github.com/qiliuchn/training-loss-pattern}.}




\maketitle

\section{Introduction}
In modern deep learning, gradient-based optimization is essential for training large-scale models efficiently. Among the most widely used optimizers is Adam, which combines the advantages of momentum and adaptive learning rates. While Adam is known for its superior convergence properties, particularly in handling noisy gradients, it sometimes exhibits a characteristic loss pattern during training: a sharp drop in loss at the start of each epoch, followed by a gradual increase. We term this behavior the \textit{Epochal Sawtooth Phenomenon} (ESP).

\begin{figure}
\centering
\includegraphics[width=1.0\linewidth]{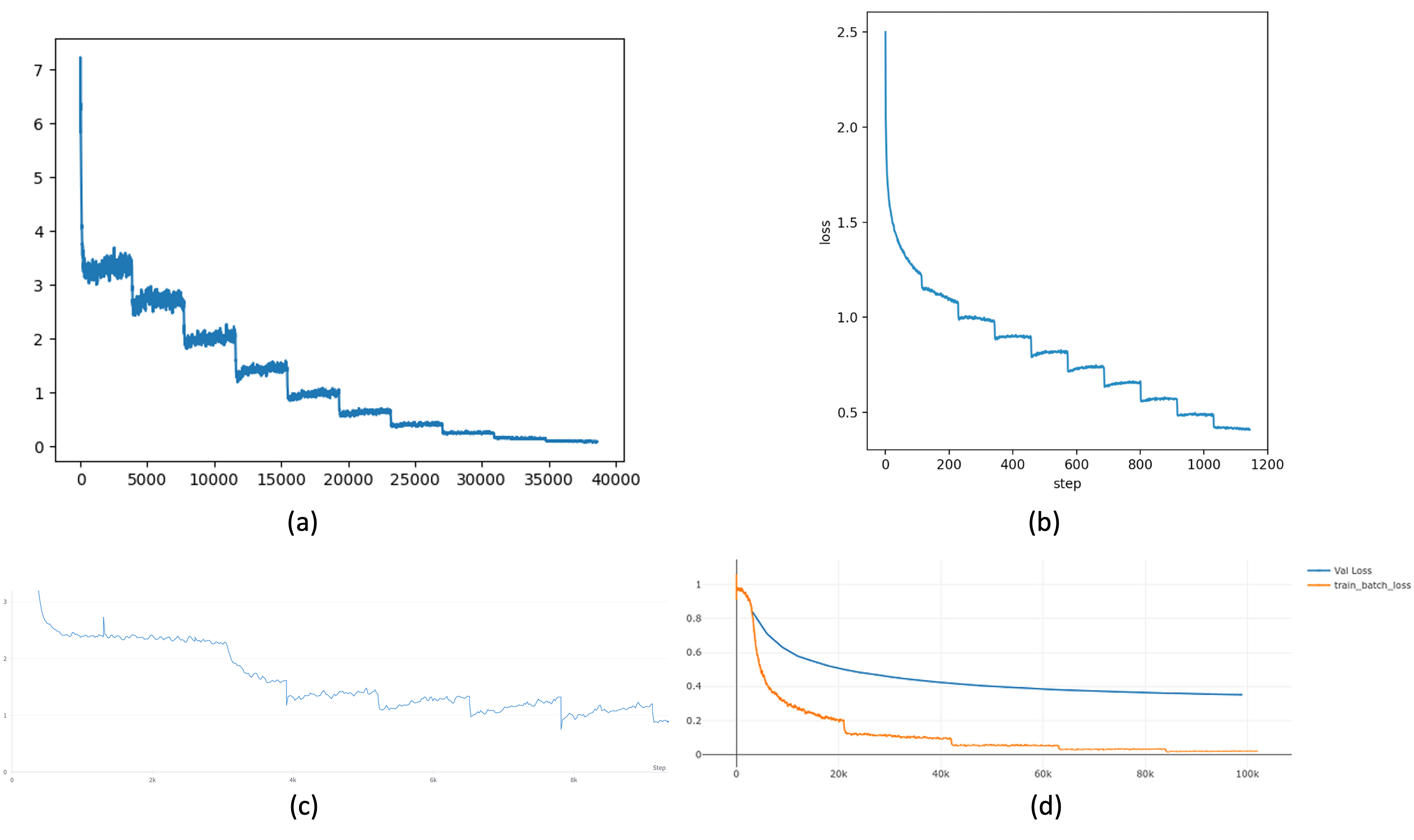}
\caption{\label{fig:intro-example}Training loss sawtooth pattern observed by practitioners; (a) A Huggingface Transformers model \cite{huggingface2025steploss}; (b) A CLIP model \cite{huggingface2025transformersissue}; (c) 
A CNN model \cite{pytorch2025trainingloss}; (d) A Classification Model \cite{pytorch2025losscurve}.
}
\end{figure}

Although this phenomenon is especially pronounced when using Adam, it is not exclusive to this optimizer. Variants of the pattern can be observed in other optimizers, such as RMSProp, though the severity is typically reduced. Often overlooked or attributed solely to data reshuffling at the beginning of each epoch, the sawtooth pattern actually reflects deeper dynamics in how optimizers adjust step sizes and learning rates across gradients of varying magnitudes. This behavior has been noted by many practitioners, as illustrated in Figure \ref{fig:intro-example}, with concerns about whether it signals overfitting.

The primary aim of this paper is to provide a comprehensive explanation and quantitative analysis of the Epochal Sawtooth Phenomenon. In particular, we focus on how Adam’s parameter settings, especially $\beta_1$ and $\beta_2$, influence the appearance of this pattern. While $\beta_1$, which controls the momentum term, plays a relatively smaller role, $\beta_2$ has a significant impact on the shape of the loss curve and training stability, affecting how the optimizer adapts to gradients of varying magnitudes. Higher values of $\beta_2$ can lead to a more gradual, near-linear increase in loss after the initial drop, while lower values of $\beta_2$ result in a sharp, concave upward trend. The common interpretation that the optimizer struggles to adapt to harder examples in each epoch is misleading, as the samples have been randomly shuffled.

To quantitatively analyze this behavior, we replicate the Epochal Sawtooth Phenomenon using a controlled quadratic minimization problem. By solving high-dimensional quadratic optimization tasks incrementally, we are able to observe this phenomenon even in simplified settings, demonstrating its generality across different types of optimization problems. Through this controlled setup, we highlight how both adaptive learning rates and data reshuffling at the start of each epoch contribute to the formation of ESP. Additionally, we show that not shuffling the data can significantly reduce ESP, while reversing the sample sequence for each epoch can amplify ESP. We also attempt to provide explanations for these observations.

The contributions of this paper are:
\begin{itemize}
    \item We identify a common loss pattern observed during training with adaptive gradient-based optimizers. Specifically, we highlight the factors influencing the magnitude of this pattern, including data shuffling, data replacement, Adam $\beta$ settings, batch size, and model capacity.
    \item We provide an in-depth explanation of the underlying mechanisms that lead to ESP. We find that \textquotedblleft immediate re-exposure to samples \textquotedblright effects means that the model will learn more (or memorize more) at the beginning of a epoch. Smaller $\beta_2$ exacerbate ESP and can act as a form of regularization.
    \item We replicate this pattern using high-dimensional incremental quadratic optimization, demonstrating the generality of this phenomenon.
\end{itemize}

The rest of this paper is organized as follows. In Section 2, we review related work on gradient-based optimization and loss patterns during training. Section 3 analyzes various factors and their effects on Epochal Sawtooth Phenomenon, and provides a detailed analysis of Adam's update rules and explains how $\beta$ settings affect the loss dynamics. In Section 4, we present a quantitative study using quadratic minimization to replicate ESP. Finally, we conclude in Section 5 with potential future directions for research on optimizing training dynamics in gradient-based optimization.

\section{Related Work}
The field of gradient-based optimization has seen extensive developments over the years, particularly in the context of training deep neural networks. Several methods have emerged to address the inherent challenges of non-convex optimization, high-dimensional parameter spaces, and noisy gradients. In this section, we review related work on gradient-based optimization, focusing on adaptive methods like Adam, and examine studies related to observed loss patterns during training, including phenomena akin to the Epochal Sawtooth Phenomenon.

\vspace{1em}\noindent\textbf{Gradient Descent and Its Variants} Gradient descent is a foundational optimization method that computes the gradient of a loss function to iteratively update model parameters. \textit{Stochastic Gradient Descent (SGD)} has become one of the most widely used algorithms for training deep neural networks, where the gradient is calculated using random batches of data. However, SGD can suffer from high variance in the gradient estimates, leading to unstable updates and slow convergence, particularly when the learning rate is not tuned properly \cite{ruder2016overview}.

To alleviate some of these challenges, momentum-based methods were developed. Polyak introduced the idea of using an exponentially weighted moving average of past gradients to smooth the updates and accelerate convergence \cite{polyak1964some}. This method reduces oscillations during training, particularly in high-curvature regions of the loss landscape. Nesterov Accelerated Gradient (NAG) further improved momentum by anticipating the next update, which helps to reduce overshooting when moving towards a minimum \cite{nesterov1983method}.

\vspace{1em}\noindent\textbf{Adaptive Optimization Methods} While momentum methods improved the stability of gradient descent, adaptive learning rate methods were developed to handle different parameter scales effectively. \textit{AdaGrad} introduced the idea of adjusting the learning rate for each parameter individually based on the historical sum of gradients, which made it well-suited for sparse data \cite{duchi2011adaptive}. However, AdaGrad's per-parameter learning rate tends to shrink over time, leading to vanishing updates.

To address this issue, \textit{RMSProp} was proposed to keep a running average of the squared gradients and adjust the learning rate adaptively, ensuring more balanced updates even in later stages of training \cite{tieleman2012rmsprop}. However, the most significant advancement in adaptive optimization came with \textit{Adam} (Adaptive Moment Estimation), which combines both momentum and adaptive learning rates \cite{kingma2014adam}. Adam maintains exponentially decayed averages of both the first moment (mean of the gradients) and the second moment (variance of the gradients), controlled by the parameters $\beta_1$ and $\beta_2$, respectively. Adam's ability to handle noisy gradients and adapt learning rates dynamically has made it one of the most popular optimizers in deep learning.

\vspace{1em}\noindent\textbf{Loss Patterns During Training} Smith et al. \cite{smith2017cyclical} observed similar oscillations in the loss function during experiments with cyclical learning rates (CLR). In their work, they proposed periodically varying the learning rate to improve training performance by capitalizing on these oscillatory behaviors. By allowing the learning rate to fluctuate, they were able to escape sharp minima and improve generalization. Stochastic Weight Averaging (SWA) further exploits these oscillations by averaging the weights of a model over multiple points during training, smoothing the oscillations in the loss function and achieving more stable convergence \cite{izmailov2018averaging}. Smith et al. \cite{smith2019super} introduced the idea of using cyclical learning rates (CLR), where the learning rate oscillates between a minimum and maximum bound during training. This schedule was found to help models escape sharp minima, improve generalization, and speed up convergence, especially in deep convolutional networks. CLR has also been integrated with other optimization techniques, such as Adam, showing that adaptive learning rates combined with periodic cycling can enhance performance across a range of tasks \cite{smith2017cyclical}. Understanding the dynamics of adaptive optimizers like Adam is crucial for improving their performance in deep learning tasks. A deeper theoretical understanding of these optimizers was provided by Wilson et al. \cite{wilson2017marginal}, who examined how the first and second moment estimates used in Adam influence the trajectory of the loss function during training. Their work demonstrated that the optimizer's dynamics can lead to unexpected loss patterns, such as oscillations, depending on the choice of hyperparameters. 

\vspace{1em}\noindent\textbf{The Role of $\beta_1$ and $\beta_2$ in Adam} The parameters $\beta_1$ and $\beta_2$ in Adam control how the optimizer adapts to gradients during training. Kingma showed that $\beta_1$, which controls the momentum term, affects how rapidly the optimizer responds to the gradient direction  \cite{kingma2014adam}. A larger $\beta_1$ results in smoother updates over time by considering past gradients, while smaller $\beta_1$ makes the optimizer more reactive to recent gradients. The second moment estimate, controlled by $\beta_2$, plays a more dominant role in determining how Adam adjusts the learning rate based on the variance of the gradients. When $\beta_2$ is close to 1, Adam smooths the second moment estimate over a long period, resulting in stable, linear loss behavior, especially during the increasing phases of training \cite{kingma2014adam, loshchilov2017decoupled}. On the other hand, when $\beta_2$ is small, the optimizer reacts more quickly to changes in gradient magnitudes, resulting in concave loss patterns, as the learning rate adapts more aggressively to the current batch of data \cite{loshchilov2017decoupled}.

\vspace{1em}\noindent\textbf{Quadratic Minimization for Understanding Optimizer Dynamics} Quadratic minimization problems have often been used as a theoretical framework to understand optimizer dynamics in machine learning \cite{Goodfellow16deep}. Bottou and Bousquet demonstrated that quadratic loss functions allow for more precise analysis of the stability and convergence properties of different optimization algorithms \cite{bottou2007tradeoffs}. By studying quadratic problems, researchers can isolate specific behaviors of optimizers, such as step size adaptation and gradient smoothing \cite{bottou2010large}.

\vspace{1em}\noindent Much of the research in gradient-based optimization has focused on improving convergence and efficiency. However, the periodic loss oscillations observed during training, especially with adaptive optimizers like Adam, highlight the complexity of the optimization process. Our work contributes to this understanding by providing a quantitative analysis of how $\beta_1$ and $\beta_2$ and many other factors affect the loss dynamics and by replicating the phenomenon in controlled quadratic minimization tasks. The Epochal Sawtooth Phenomenon thus provides a new lens through which to examine the nuances of gradient-based optimization.

\section{Empirical Study}
Firstly, we replicate the Epochal Sawtooth Phenomenon by pre-training BERT models. Two variants are used: BERT-small, which consists of 768 hidden units, 6 encoder layers, and 6 attention heads, and BERT-tiny, which has 256 hidden units, 2 encoder layers, and 2 attention heads. These models are similar to, but not exact replicas of, the Google-published models (\cite{devlin2019bert}) with the same names. The code for these models is shared at \cite{liu2023trainingloss}. BERT-small is trained from scratch on the 269MB Wikitext dataset (\cite{wikitext}), while BERT-tiny is trained on a smaller 11MB subset of the same dataset. We use \textit{Byte pair encoding (BPE)} to build a 30k-sized vocabulary instead of \textit{Wordpiece}. The benchmark settings include a batch size of 128 for BERT-small and 32 for BERT-tiny, with the DataLoader shuffling after every epoch. The standard Masked Language Model (MLM) and Next Sentence Prediction (NSP) tasks are used for pre-training. We employ the PyTorch Adam optimizer with default $\beta$ values of (0.9, 0.999). The learning rate is set to $1 \times 10^{-4} $, and the weight decay rate is $1 \times 10^{-5} $. For the purpose of this discussion and to reproduce the desired effect, we temporarily disregard the usual practice of setting the weight decay rate higher during pre-training. In this study, when we say we “shuffle” the data, we mean that the DataLoader will shuffle the samples for each epoch of training. “No shuffle” means that the dataset sample sequence remains the same across all epochs. “Sample with replacement” means that a batch of data will be randomly sampled from the dataset at each time step. “Reverse sample sequence” means that the dataset sample sequence is reversed for each epoch.

\begin{figure}
\centering
\includegraphics[width=1.0\linewidth]{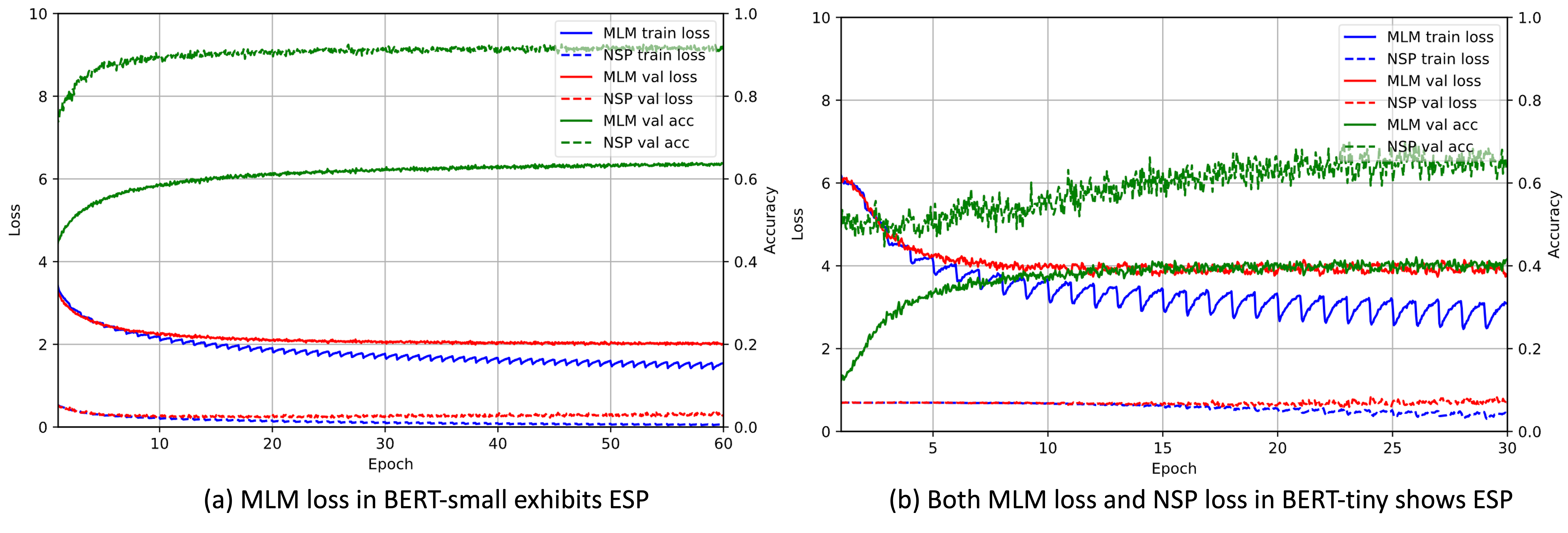}
\caption{\label{fig:reproduce} Epochal Sawtooth Phenomenon (ESP) reproduced using BERT-small and BERT-tiny; (a) MLM loss in BERT-small exhibits ESP; (b) Both MLM loss and NSP loss in BERT-tiny exhibit ESP. The training loss drops sharply at the start of each epoch; then gradually increase over the epoch, when the overall trend remains downward.
}
\end{figure}

The MLM training loss for both models reproduces the sawtooth pattern, as shown in Figure \ref{fig:reproduce} (a) and (b). The NSP loss, which is much smaller in magnitude, also exhibits this pattern, particularly with smaller $\beta_2$, as demonstrated in Figure \ref{fig:reproduce} (b). The losses are averaged over 500 batches for BERT-small and 50 batches for BERT-tiny, without smoothing. As seen in the figure, the training loss drops sharply at the beginning of each epoch and then gradually increases throughout the epoch. However, the overall trend remains downward.

\subsection{Influencing Factors}

\begin{figure}
\centering
\includegraphics[width=1.0\linewidth]{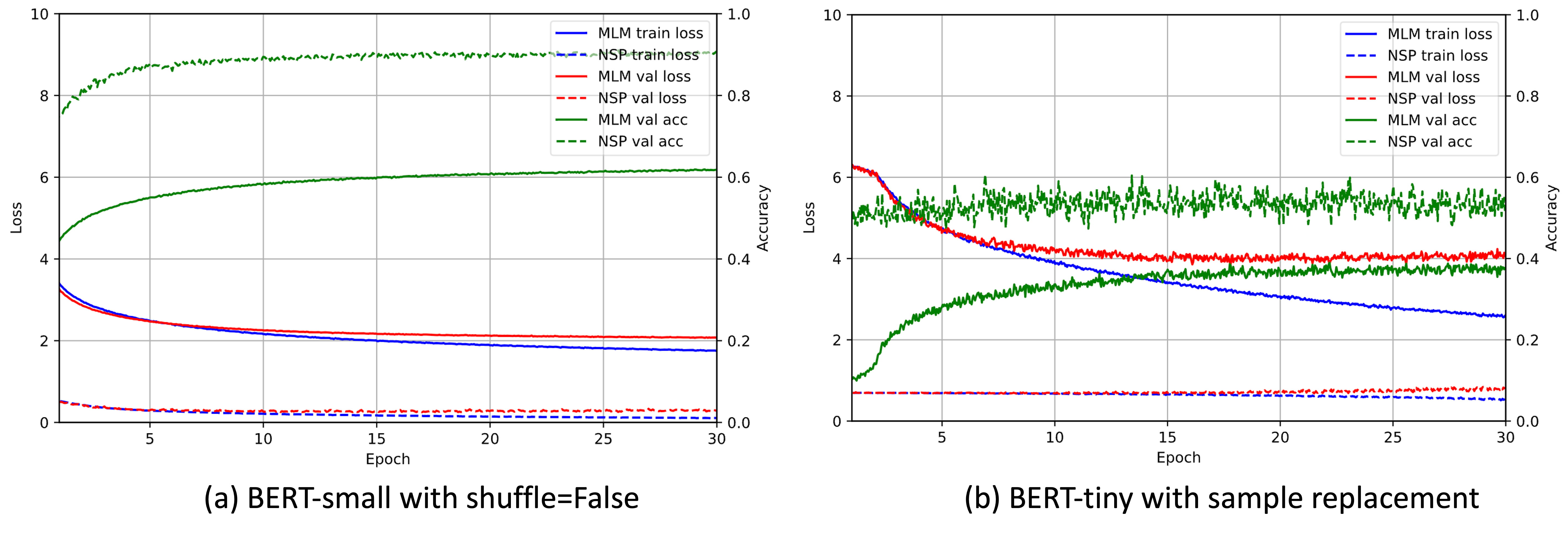}
\caption{\label{fig:shuffle-replacement} Effects of sample shuffling and replacement settings on ESP: (a) BERT-small with \texttt{shuffle=False}; (b) BERT-tiny with sample replacement. ESP nearly vanishes when the dataloader does not shuffle the data, and it disappears completely when replacement is used.
}
\end{figure}

In this section, we examine the factors influencing ESP. We primarily use BERT-small for illustration, though many other models exhibit similar results. First, we observe that this phenomenon nearly vanishes (Figures \ref{fig:shuffle-replacement}(a) and (b)), with only a slight trace remaining, when the dataloaders do not shuffle at the start of each epoch (Figure \ref{fig:shuffle-replacement}(a)). It disappears entirely when sampling with replacement for each batch. In this case, the concept of an ``epoch'' becomes irrelevant. This observation suggests that the Epochal Sawtooth Phenomenon (ESP) is closely tied to the sequencing of samples.

\begin{figure}
\centering
\includegraphics[width=1.0\linewidth]{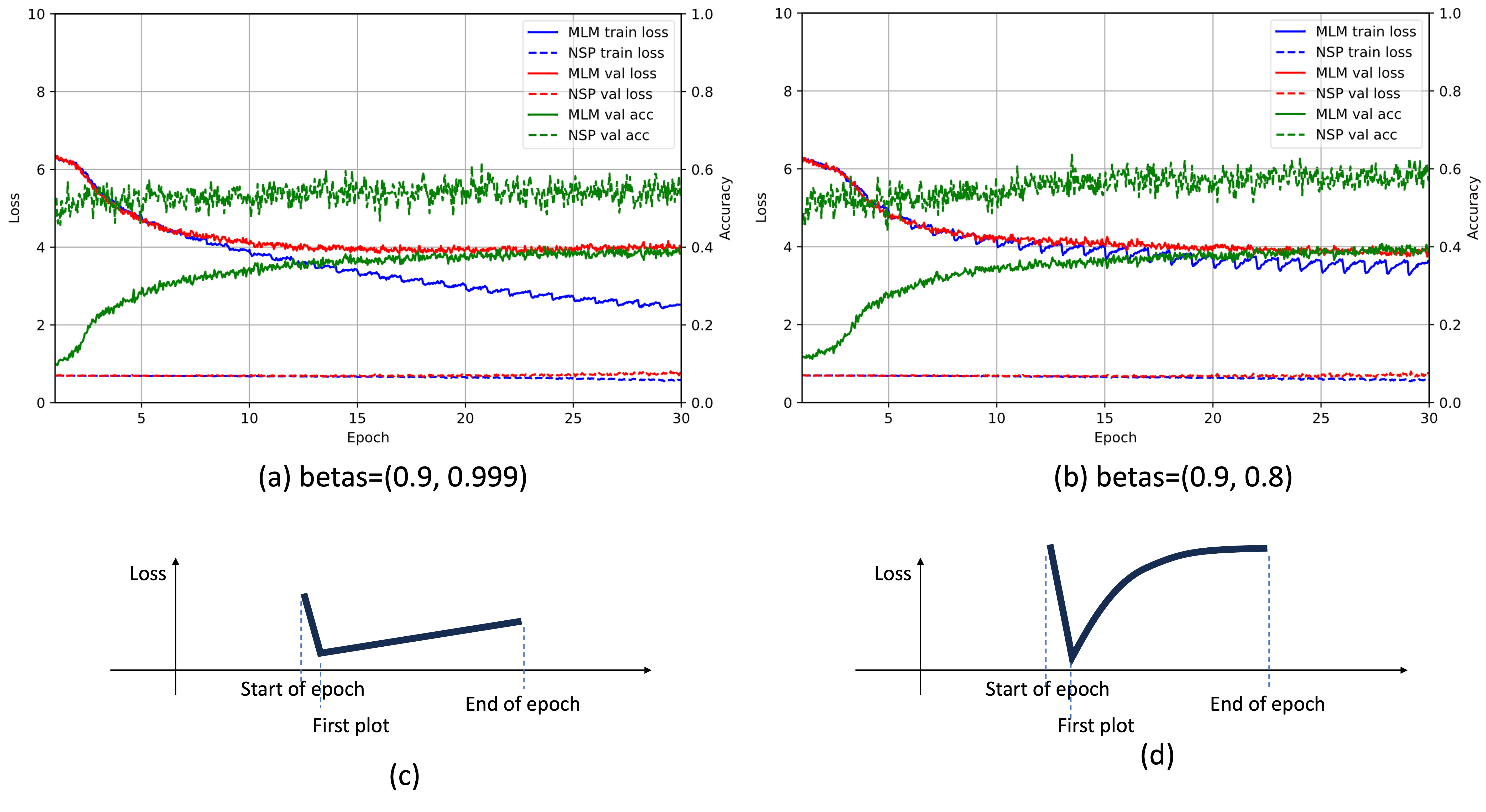}
\caption{\label{fig:beta-2} Effects of Adam $\beta_2$ setting on ESP. (a) BERT-tiny with \texttt{betas=(0.9, 0.999)}; (b) BERT-tiny with \texttt{betas=(0.9, 0.8)}. As $\beta_2$ gets smaller, ESP is amplified; the increasing part of the curve goes from being seemingly linear to obviously concave.
}
\end{figure}

We first explore the effect of Adam’s $\beta_2$ parameter, as shown in Figure \ref{fig:beta-2}. It is clear that as $\beta_2$ decreases, the ESP is amplified, as shown by Figure \ref{fig:beta-2}. Interestingly, as $\beta_2$ gets smaller, the increasing part of the curve transitions from being seemingly linear (Figure \ref{fig:beta-2}(a)) to distinctly concave (Figure \ref{fig:beta-2}(b)). We note that $\beta_1$ does not have a noticeable effect on ESP, although smaller values of $\beta_1$ can make the ESP pattern less discernible due to higher variance (See additional figures on Github).

The weight decay parameter has a relatively small impact on ESP. Larger values of \texttt{weight\_decay} slightly weaken the ESP. Model size, however, has a significant impact on ESP. As the model size increases, ESP becomes more prominent. It is worth considering whether ESP is a clear indicator of over-fitting; if so, reducing model capacity would be preferable (See additional figures on Github).

\begin{figure}
\centering
\includegraphics[width=1.0\linewidth]{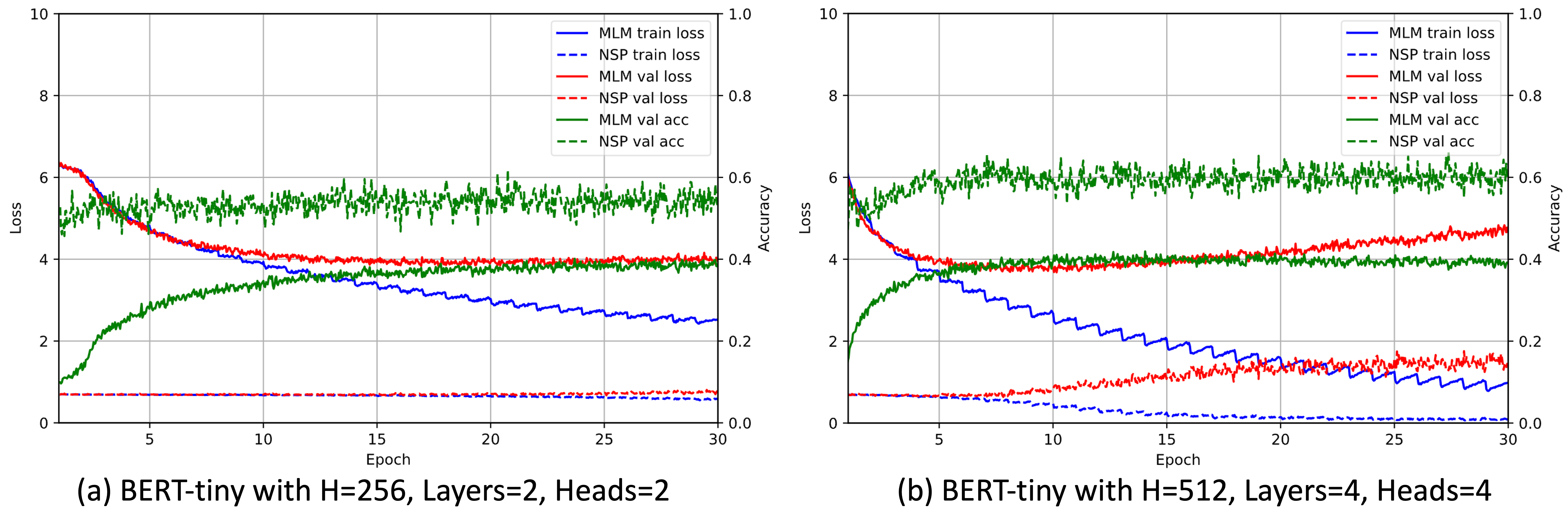}
\caption{\label{fig:model-size} Effects of model size settings on ESP; (a) BERT-tiny with \texttt{H=128, Layers=2, Heads=2}; (b) BERT-tiny with \texttt{H=512, Layers=4, Heads=4}. Model capacity has a substantial impact on ESP; larger model sizes exacerbate ESP.
}
\end{figure}

To summarize:
\begin{itemize}
\item No shuffling the data can significantly reduce ESP; using sample replacement can completely eliminate ESP.
\item Adam parameter $\beta_2$ has a major impact on ESP; a smaller $\beta_2$ results in more pronounced ESP. $\beta_1$ does not have a noticeable effect on ESP.
\item Model capacity has a substantial impact on ESP; larger model sizes exacerbate ESP.
\item A higher weight decay rate slightly weakens ESP.
\end{itemize}

\subsection{Interpretation and Analysis}
\subsubsection*{Notation List}
\begin{description}
    \item[$e$] Epoch index;
    \item[$t$] Step count for a specific epoch;
    \item[$N$] Total number of samples;
    \item[$B$] Batch size; $b$ is used for indexing;
    \item[$l^b_t$] Loss of batch $b$ at step $t$ (epoch index is ignored);
    \item[$l_0$] Expected loss of the batch at the start of an epoch (epoch index is ignored);
    \item[$\nabla l^b_t$] Gradient of the loss for batch $b$ at step $t$;
    \item[$g_t := \nabla l^t_t$] Gradient of the loss for batch $t$ at step $t$; this is the gradient used for updating $\theta$;
    \item[$m_t$] Adam's first momentum at step $t$; $\hat{m_t}$ is $m_t$ with initialization correction;
    \item[$v_t$] Adam's second momentum at step $t$;
    \item[$\Delta \theta_t$] Update to $\theta$ at step $t$.
\end{description}

\begin{figure}
\centering
\includegraphics[width=1.0\linewidth]{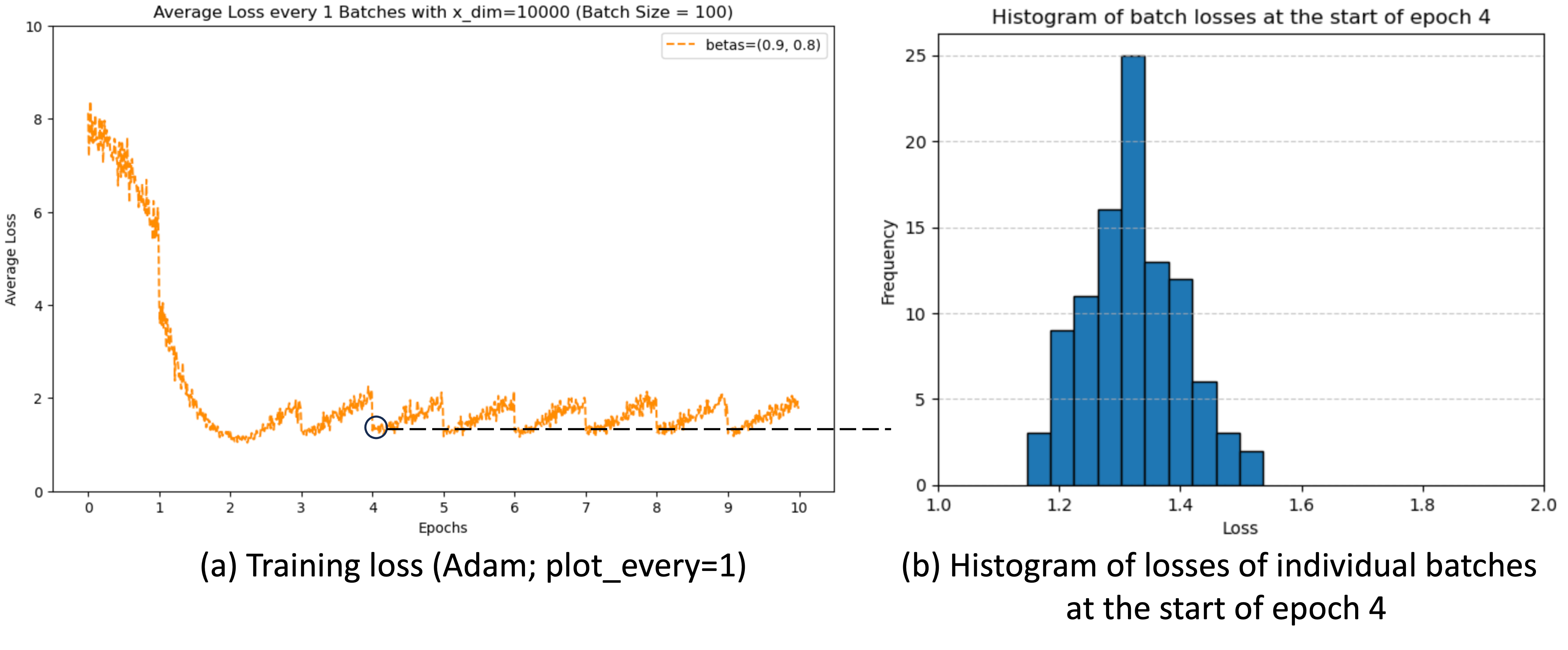}
\caption{(a) Training loss curve when \texttt{plot\_every=1}. (b) Histogram of loss of all batches at the start of epoch 4.
}
\label{fig:loss-all-batches-init}
\end{figure}

\begin{figure}
\centering
\includegraphics[width=1.0\linewidth]{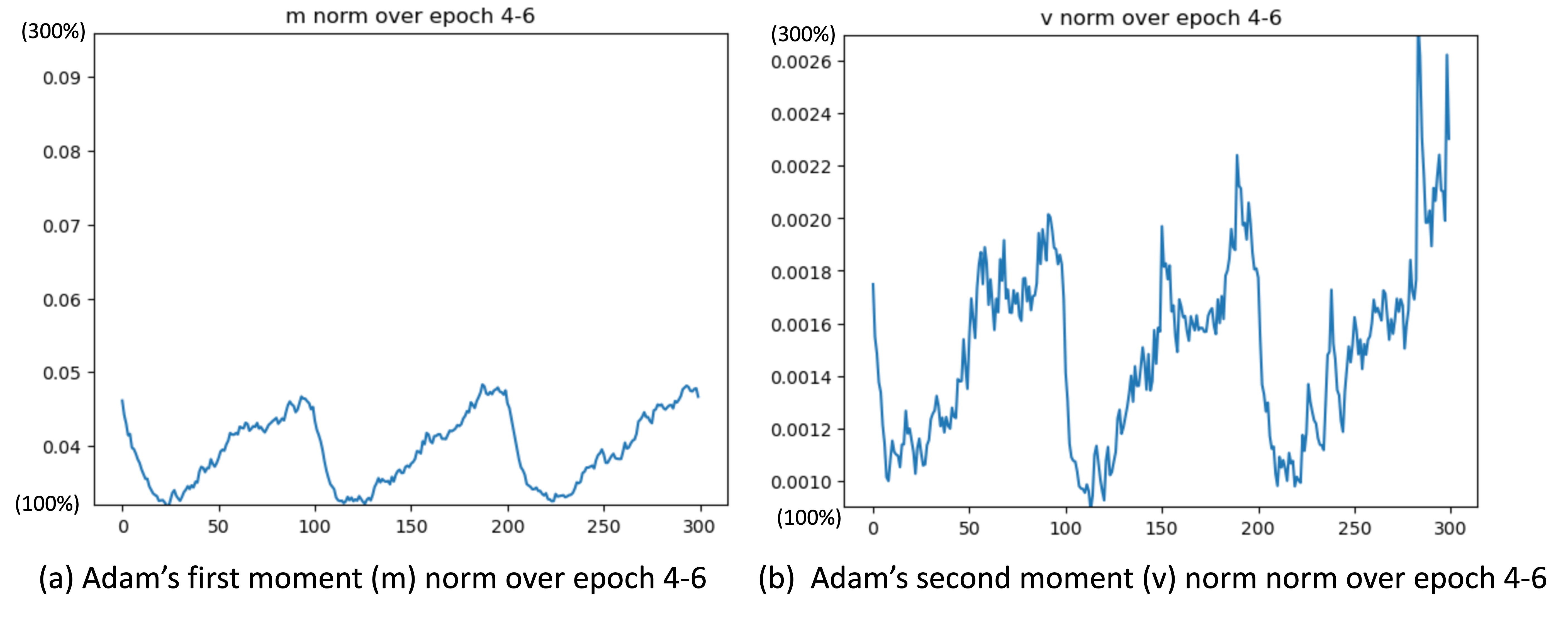}
\caption{(a) $m$ norm over epoch 4-6; (b) $v$ norm over epoch 4-6. Momentum takes on large value at the beginning of epoch, then drop exponentially; afterwards gradually increases. $\|v\|$ steadily increases during epoch.
}
\label{fig:m-v-norms}
\end{figure}

Suppose there are $N$ samples in total, and the batch size is $B$. The data loader shuffles the data for each epoch, and samples are drawn without replacement. All batches have the same distribution of loss values at the beginning of the epoch due to shuffling, as shown in Figure \ref{fig:loss-all-batches-init} (b). Without loss of generality, we assume that the loss values are normally distributed with an expected value of $l_0$ and variance $\frac{\sigma^2}{B}$. The last few batches from epoch $e - 1$ and the first few batches from epoch $e$ will share some samples in common. For example, the last batch of epoch $e - 1$ and the first batch of epoch $e$ are expected to have $\frac{B^2}{N}$ samples in common. Hence, the momentum is expected to be large at the beginning of an epoch, as shown in Figure \ref{fig:m-v-norms} (a). This momentum is much larger compared to when we do not shuffle, in which case each sample must wait for exactly one epoch before being seen again.

We list the Adam update rules (Eqn \ref{eq:adam-m}, \ref{eq:adam-v}, \ref{eq:adam-theta}) here for clarity (weight decay and initialization corrections are omitted for simplicity). A simple example can illustrate how sample shuffling and the introduction of momentum can cause the training loss curve to oscillate. Let the total loss be $f(\theta) = g(\theta) + h(\theta)$, where $g(\theta) = \theta$ and $h(\theta) = 1 - \theta$. Suppose we try to ``optimize'' $f$ incrementally, where \(g\) is the loss of the first batch and \(h\) is the loss of the second batch. The total loss \(f\) is constant, but the individual batch losses are not.

\begin{figure}
\centering
\includegraphics[width=1.0\linewidth]{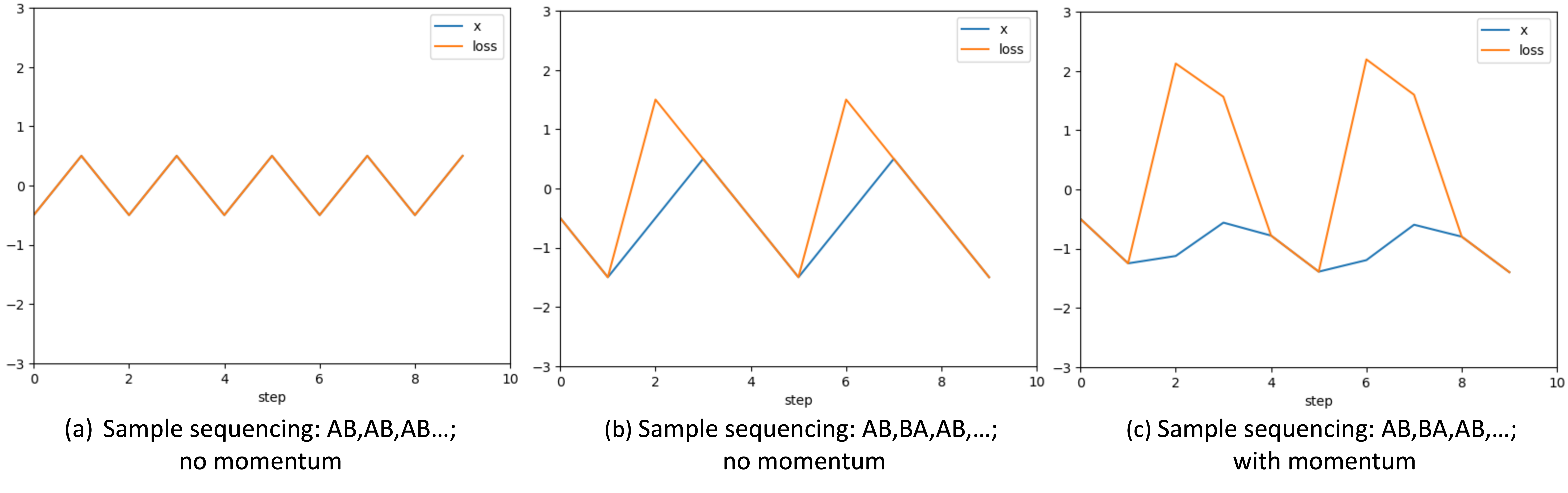}
\caption{Illustration of how sample sequencing and momentum can cause training loss oscillations. (a) sample sequencing: AB, AB..., mimicking no shuffling; (b) sample sequencing: AB, BA, AB,...., reversing sample sequence for each epoch; this strategy exaggerates the effect of shuffling; (c) momentum is added. Batch loss oscillates more when shuffling and momentum exists.
}
\label{fig:sequence-momentum-illust}
\end{figure}

First, suppose that we sequence the samples as AB, AB, AB…, mimicking the scenario with no shuffling, and we use the gradient method to update $\theta$. The loss curve is shown in Figure \ref{fig:sequence-momentum-illust} (a). We can observe that the loss curve oscillates slightly. For the second scenario, we sequence the samples as AB, BA, AB…, exaggerating the effect of sample shuffling. The last sample from the previous epoch is seen again immediately when the new epoch starts. We can observe that the batch loss curve oscillates more, as shown in Figure \ref{fig:sequence-momentum-illust} (b). Next, we add momentum, and we can see that the batch loss curve oscillates even more, as illustrated in Figure \ref{fig:sequence-momentum-illust} (c).

\begin{equation}
    \label{eq:adam-m}
    m_t = \beta_1 m_{t-1} + (1 - \beta_1) g_t
\end{equation}

\begin{equation}
    \label{eq:adam-v}
    v_t = \beta_2 v_{t-1} + (1 - \beta_2) g_t^2
\end{equation}

\begin{equation}
    \label{eq:adam-theta}
    \theta_t = \theta_{t-1}  - \gamma \frac{m_t}{\sqrt{v_t} + \epsilon}
\end{equation}

\begin{figure}
\centering
\includegraphics[width=1.0\linewidth]{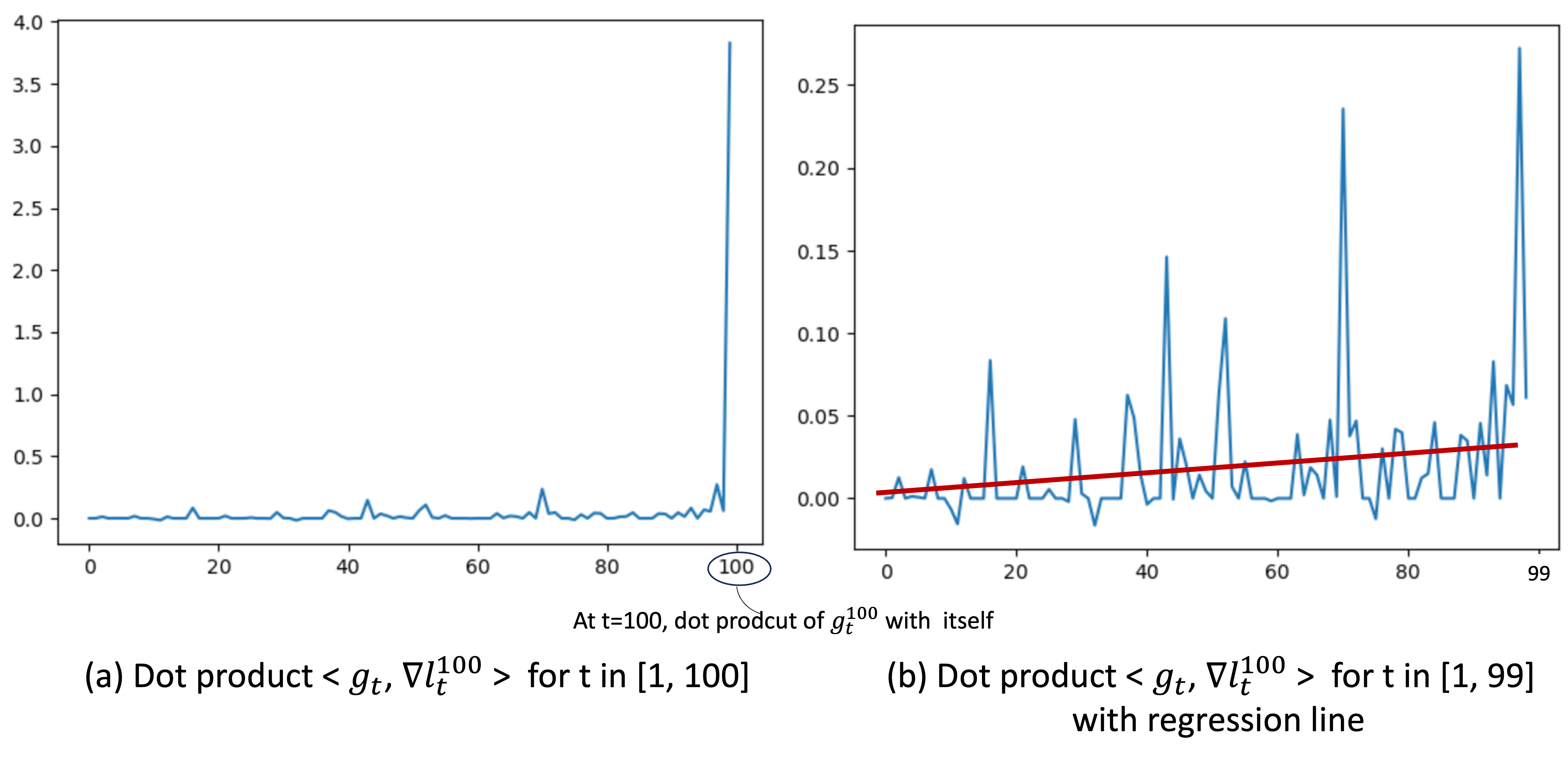}
\caption{$\langle g_t,\nabla l^b_t \rangle$ for batch \texttt{b=100}. (a) $t$ is in range $[1, 100]$. (b) $t$ is in range $[1, 99]$. The product is largest when $t=100$ since $g_{100} = \nabla l^{100}_{100}$. For clarity, we separate $t$ in $[1,99]$ from $t=100$ case. The dot products are mostly positive, which means learning batch $t$'s gradient loosely aligns with gradient of batch $b$ (angle slightly smaller than $\pi/2$). The gradient is slowly increasing due to the fact $\theta$ is getting gradually away from the optima of batch $b$.
}
\label{fig:dot-gt-g100}
\end{figure}

\begin{figure}
\centering
\includegraphics[width=1.0\linewidth]{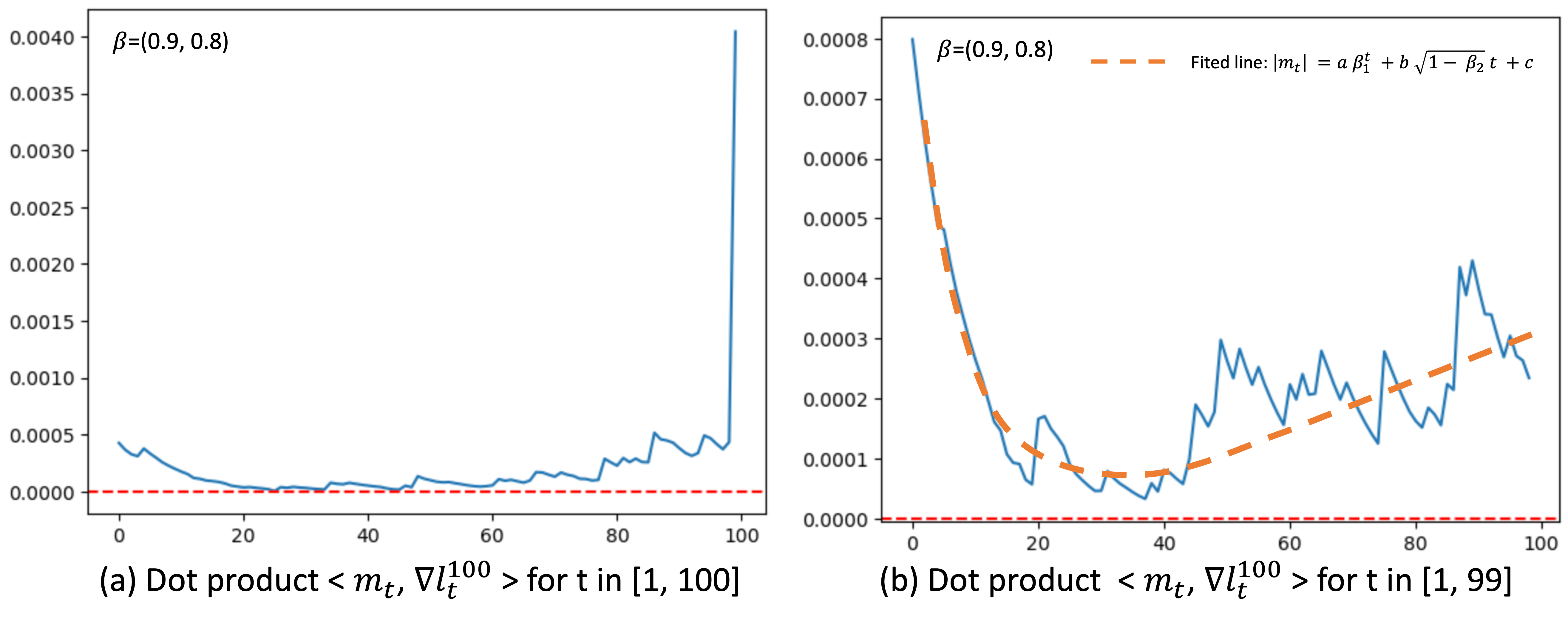}
\caption{(a) \(\langle m_t, \nabla l^b_t \rangle\) over steps $t$ in [1, 100] for \texttt{b=100}. (b) \(\langle m_t, \nabla l^b_t \rangle\) over steps $t$ in [1, 99] for \texttt{b=100}.
}
\label{fig:dot-mt-g100}
\end{figure}

Let $l^b_t$ denote the loss of batch $b$ at step $t$ of epoch $e$ (with the index $e$ ignored). Let step $t$ be 0 at the start of epoch $e$. At each step $t$, batch $t$'s gradient $\nabla l^t_t$ will be used to update the model parameters $\theta$. By convention, we denote this as $g_t := \nabla l^t_t$. We fix a batch $b$ and analyze how its loss changes over the epoch, especially for $t < b$. Through experiments, we find that $\langle g_t, \nabla l^b_t \rangle$ is mostly a small positive value for $t < b$, which means batch $t$'s gradient loosely aligns with the gradient of batch $b$. In other words, their angle is slightly less than $\pi / 2$. Of course, $\langle g_t, \nabla l^b_t \rangle$ is greatest when $t = b$ since $g_b = \nabla l^b_b$. This is also true for the momentum. These observations are shown in Figure \ref{fig:dot-gt-g100} and Figure \ref{fig:dot-mt-g100}. The norm $\|g_t\|$ is approximated by Eqn \ref{eq:grad-norm}, where $a_g$ and $b_g$ are fitting parameters. 

\begin{figure}
\centering
\includegraphics[width=1.0\linewidth]{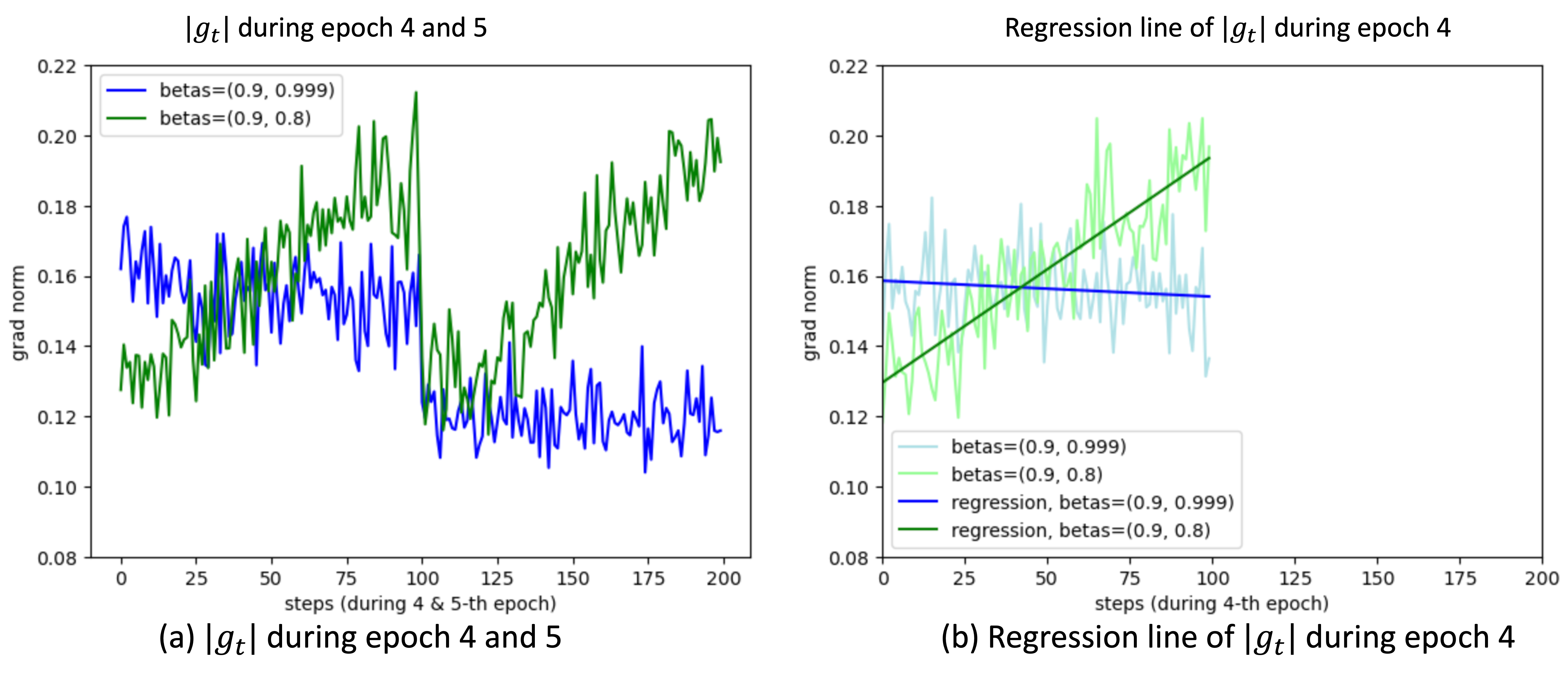}
\caption{(a) $\|g_t\|$ over epoch 4 and 5; (b) Regression line of $\|g_t\|$ over epoch 4. $\|g_t\|$ restart from a relatively small value at the beginning of each epoch and steadily increases over the epoch if $\beta_2 < 0.9$.
}
\label{fig:grad-norm}
\end{figure}

The increase in $\|g_t\|$ can be explained by the parameter moving away from the optima. This ``moving-away'' happens because, although $g_t$ aligns with $\nabla l^b_t$, $\theta$'s update, $\Delta \theta_t$, does not necessarily point towards reducing $l^b$. This is the reason why we observe $l^b$ increasing during the epoch. The gradient will be larger when further away from the optima, which explains the steady increase of $\|g_t\|$ over the course of an epoch (see Figure \ref{fig:grad-norm}).

\begin{equation}
    \label{eq:grad-norm}
    \|g_t\| \approx a_g  + b_g \sqrt{1 - \beta_2} t
\end{equation}

Momentum is the exponential averaging of gradients. As we mentioned earlier, large momentum arises from the immediate re-exposure to some samples at the beginning of an epoch. See Figure \ref{fig:m-v-norms} (a) for the momentum norm $\|m\|$ curve. The norm $\|m\|$ is large at the start of an epoch, then rapidly decreases, and eventually increases linearly until the end of the epoch. The exponential drop in momentum occurs as the ``immediate re-exposure to samples'' effect weakens. Afterward, momentum gradually increases because the norm $\|g_t\|$ is steadily increasing (see Eqn \ref{eq:grad-norm}). The norm $\|m\|$ is approximated by Eqn \ref{eq:m-norm}, where $a_m$, $b_m$, and $c_m$ are fitting parameters. 

The norm $\|v\|$ is expected to increase quadratically (see Eqn \ref{eq:v-norm}) as $g_t$ increases linearly. The model takes large steps toward minimizing the losses of the first few batches. In simpler terms, the model memorizes the first few batches better than the others. This is not necessarily a problem, since the samples are reshuffled each epoch. The dot product between momentum $m_t$ and $\nabla l^b_t$ is shown in Figure \ref{fig:dot-mt-g100}. It exhibits an exponential decreasing part followed by a linear increasing part. This pattern can be well approximated by Eqn \ref{eq:dot-m-grad}, where $a_{mg}$, $b_{mg}$, and $c_{mg}$ are fitting parameters.

\begin{equation}
    \label{eq:m-norm}
    \|m_t\| \approx  a_m \beta_1^t + b_m \sqrt{1 - \beta_2} t + c_m
\end{equation}

\begin{equation}
    \label{eq:v-norm}
    \|v_t\| \approx a_v + b_v t + c_v (1 - \beta_2) t^2
\end{equation}

\begin{equation}
    \label{eq:dot-m-grad}
    \begin{split}
    \langle m_t, \nabla l^b_t\rangle \approx a_{mg} \beta_1^t + b_{mg} \sqrt {1 - \beta_2} t + c_{mg} \\ 
    a_{mg}, b_{mg} \geq 0, 0 < t < b
    \end{split}
\end{equation}

For adaptive optimizers like \texttt{Adam} and \texttt{RMSProp}, the momentum (or gradient, if there is no momentum) is divided component-wise by the second moment $v$. Dividing by $v_t$ alters the direction of $g_t$ on a small scale. This does not change its alignment with $\nabla l^b_t$ when the momentum (or gradient) is large. Let $\Delta \theta_t$ be the change of $\theta$ at step $t$. The important observation here is that when $\beta_2$ is not close to 1 (i.e., $\beta_2 < 0.85$), $\langle \Delta \theta_t, \nabla l^b_t \rangle$ is likely to be positive. This means that for $t < b$, $\theta$ will move away from the optima of $l^b$, which can be seen as a form of regularization.

\begin{figure}
\centering
\includegraphics[width=1.0\linewidth]{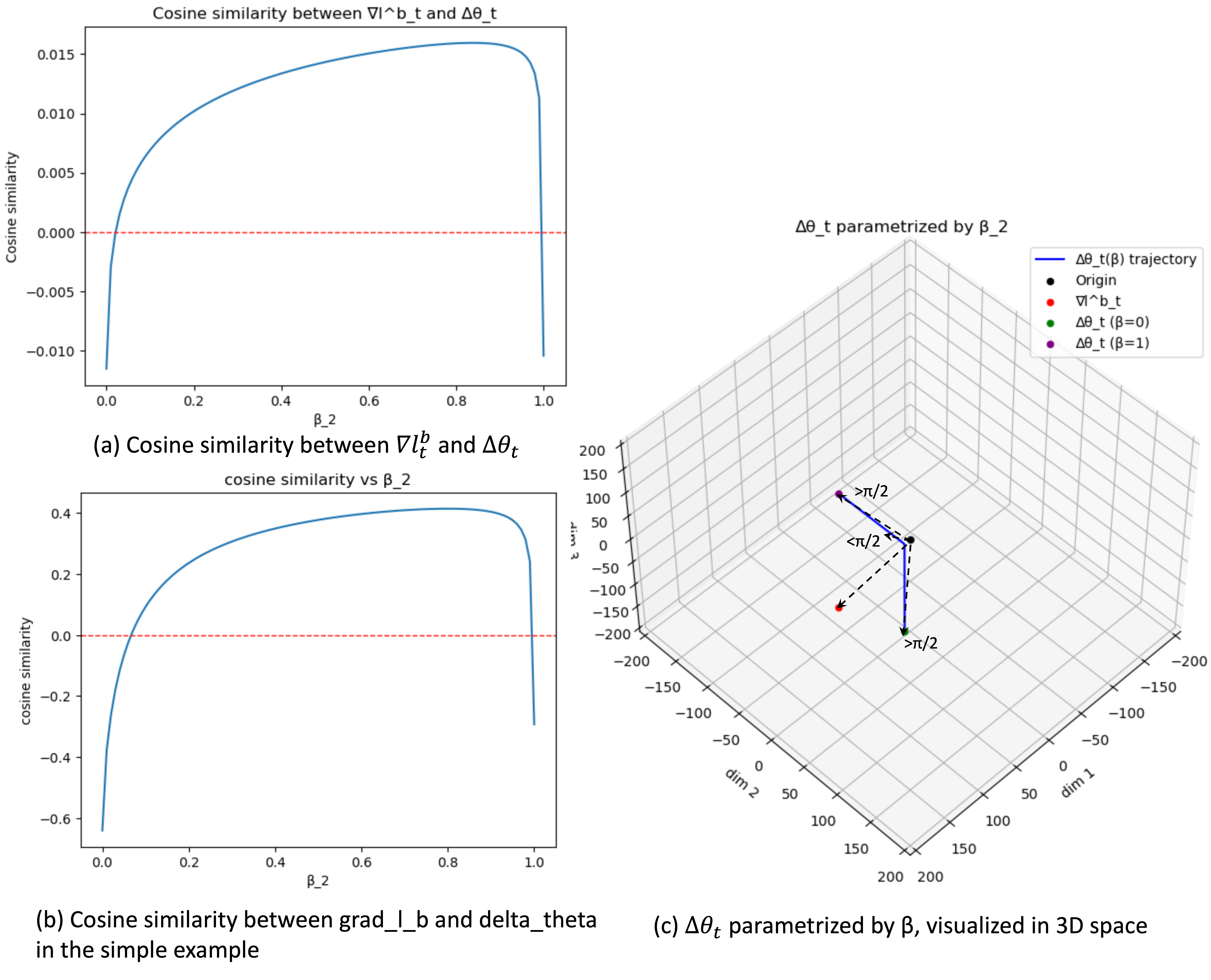}
\caption{The n-shaped similarity explained by low dimensional example; (a) Cosine similarity between $\nabla l^b_t$ and $\Delta \theta_t$; (b) 3 dimensional space example replicating the pattern in (a); (c) $\Delta \theta_t$ trajectory parametrized by $\beta_2$ visualized in 3D space illustrating cosine similarity sign change. n-shaped similarity can be explained by $m$ dividing the square root of convex combination of small and large numbers.
}
\label{fig:dot_deltax_flip_sign_explain}
\end{figure}

We also plotted the cosine similarity between $\nabla l^b_t$ and $\Delta \theta_t(\beta_2)$ as $\beta_2$ takes values in $[0, 1]$, as shown in Figure \ref{fig:dot_deltax_flip_sign_explain} (a). The cosine similarity follows an ``n''-shaped curve. $\langle \nabla l^b_t, \Delta \theta_t(\beta_2) \rangle$ is positive except when $\beta_2$ is very close to 0 or 1. A positive value of $\langle \nabla l^b_t, \Delta \theta_t(\beta_2) \rangle$ means that the update to $\theta$ will increase the batch $b$ loss $l^b$. This suggests that if $m$ component-wise divides $v_{t-1}$ or $g^2_t$, it will still align with $\nabla l^b_t$; however, if $m$ divides a convex combination, it is likely to lose alignment. This effect causes the loss to increase during the epoch, and we refer to this as the \textit{n-shaped similarity}.

To explain the n-shaped similarity, we replicate this pattern using a low-dimensional example, as shown in the code below. The three-dimensional vectors \texttt{grad\_l\_b}, \texttt{m\_hat}, \texttt{v\_prev}, \texttt{g\_squared}, and \texttt{delta\_theta} simulate $\nabla l^b_t$, $m_t$, $v_{t-1}$, $g^2_t$, and $\Delta \theta_t$, respectively. We define \texttt{v\_prev} as \texttt{[1, 0.0001, 1]} and \texttt{g\_squared} as \texttt{[1, 1, 0.0001]}. The form of $\Delta \theta_t$ is given by Eqn \ref{eq:cos-similarity}. The cosine similarity curve obtained from this simple example in 3D space is shown in Figure \ref{fig:dot_deltax_flip_sign_explain} (b). The curve for $\Delta \theta_t(\beta_2)$ in this example is shown in Figure \ref{fig:dot_deltax_flip_sign_explain} (c). The n-shaped similarity can be explained by $m$ dividing the square root of a convex combination of small (``0.0001'' in \texttt{v\_prev}, \texttt{g\_squared}) and large (``1'' in \texttt{g\_squared}, \texttt{v\_prev}) numbers, as illustrated by this simple example. Our previous analysis shows why $\beta_2$ plays a critical role in the Epochal Sawtooth Phenomenon and, more generally, in training and optimization.

\begin{equation}
    \begin{split}
        \label{eq:cos-similarity}
        \Delta \theta_t (\beta_2) = [8,  \frac{-1}{\sqrt{\beta_2*0.0001 + (1-\beta_2)*1}}, \\
        \frac{-2}{\sqrt{\beta_2*1 + (1-\beta_2)*0.0001}}]
    \end{split}
\end{equation}

\lstset{
  language=Python,
  breaklines=true,
  basicstyle=\footnotesize\ttfamily,  
  keywordstyle=\bfseries\color{blue}, 
  commentstyle=\color{commentgreen}, 
  stringstyle=\color{red},            
  showspaces=false,
  showstringspaces=false,
  frame=single,                       
  numbers=left,                       
  stepnumber=1,                       
  numbersep=5pt,                      
  numberstyle=\tiny\color{gray},      
  tabsize=2                           
}
\begin{lstlisting}
def cosine_similarity(v1, v2):
  return np.dot(v1, v2) / (np.linalg.norm(v1) * np.linalg.norm(v2))

grad_l_b = np.array([2, 1, 2])
m_hat = np.array([-8, 1, 2])
v_prev = np.array([1,0.0001,1])
g_squared = np.array([1, 1, 0.0001])

cos_sim_list = []
beta_list = np.arange(0, 1.01, 0.01)
trajectory = []
for beta2 in beta_list:
  delta_x = - m_hat / np.sqrt(beta2 * v_prev + (1 - beta2) * g_squared)
  trajectory.append(delta_x)
  cos_sim_list.append(cosine_similarity(grad_l_b, delta_x))
plt.plot(beta_list, cos_sim_list)
\end{lstlisting}

\begin{figure}
\centering
\includegraphics[width=1.0\linewidth]{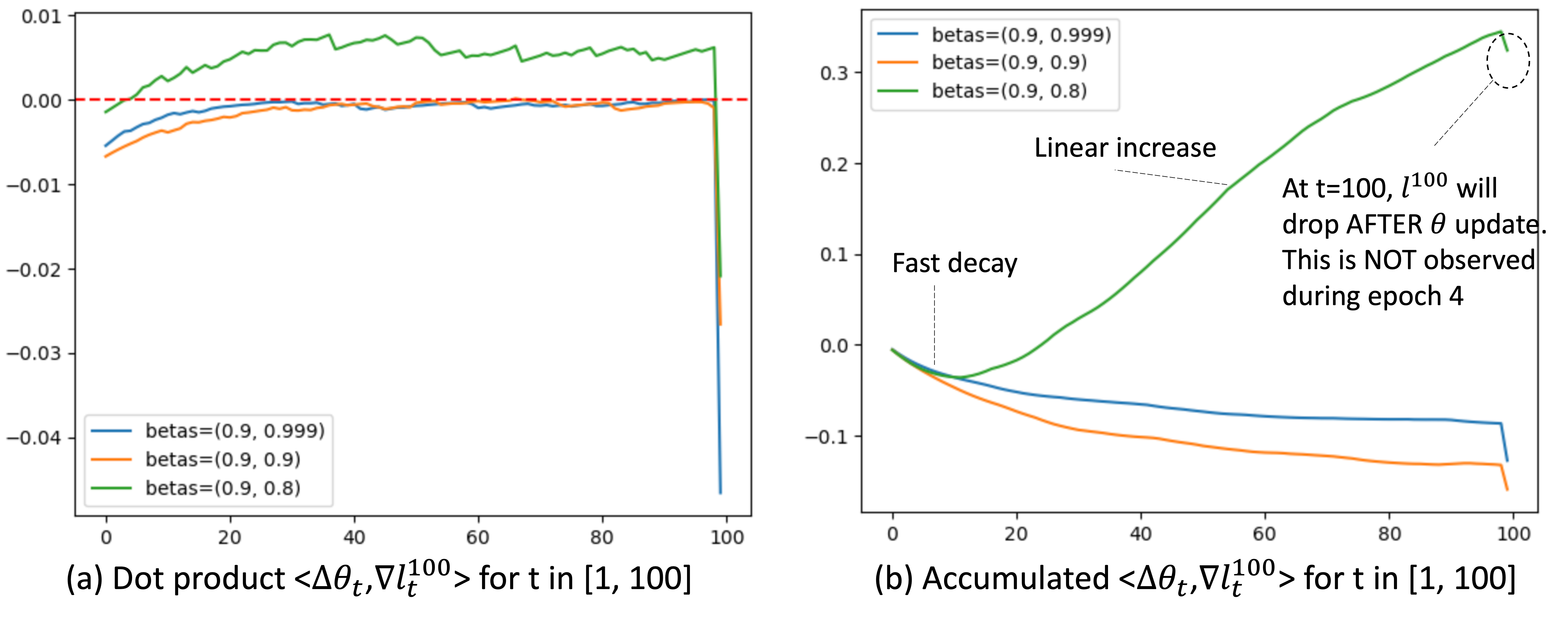}
\caption{(a) $\langle \Delta \theta_t, \nabla l^b_t \rangle$ over epoch 4 for \texttt{b=100}; (b) Accumulated $\langle \Delta \theta_t, \nabla l^b_t \rangle$ values over epoch 4 for \texttt{b=100}. The last drop happens when model is minimizing batch $b$ loss. But this drop is not observed during epoch 4.
}
\label{fig:dot-deltax-g100}
\end{figure}

\begin{figure}
\centering
\includegraphics[width=0.6\linewidth]{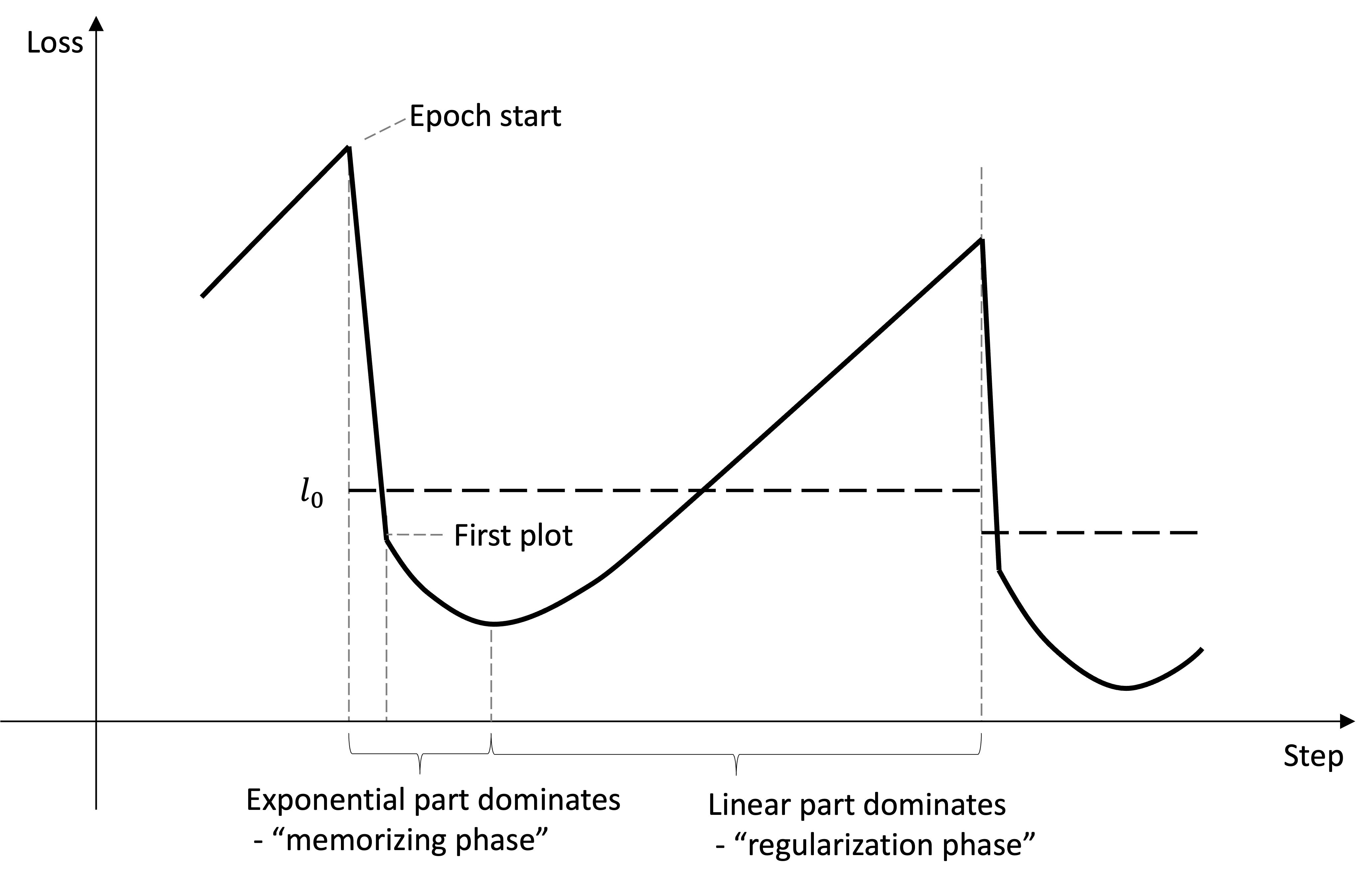}
\caption{Predicted loss curve. It can be divided into two parts. Exponential component dominates the first part; and linear component dominates the second part.
}
\label{fig:loss-final}
\end{figure}

The dot product between the update to $\theta$ ($\nabla \theta$) and the gradient of batch $b$ at step $t$, $\langle \Delta \theta_t, \nabla l^b_t \rangle$, is shown in Figure \ref{fig:dot-deltax-g100} (a). It starts from a relatively low negative value, increases exponentially towards zero, and then gradually decreases during the rest of the epoch. This behavior can be approximated by Eqn \ref{eq:dot-deltax-grad}, where $a_{xg}$, $b_{xg}$, $c_{xg}$, and $d_{xg}$ are a new set of fitting parameters. Note that Eqn \ref{eq:dot-deltax-grad} resembles Eqn \ref{eq:dot-m-grad}, but with division by $t$ and a reversal of the sign; however, the coefficients differ in this case. Batch $b$'s loss at step $t$ is approximated by Eqn \ref{eq:loss-batch-b}, and the observed training loss can be approximated by taking $b = t$. See Figure \ref{fig:loss-final} for the predicted loss curve. The curve can be divided into two phases: the ``memorizing phase'', where the exponential component dominates, and the ``regularization phase'', where the linear component dominates. The loss for each batch $b$ increases gradually when updating other batches and decreases sharply at step $b$, as illustrated by the sharp drop of batch 100 at step 100 in Figure \ref{fig:dot-deltax-g100} (b). However, this drop is only observed in the following epoch. This explains why we see a sharp drop in training loss at the start of each epoch.

\begin{equation}
    \label{eq:dot-deltax-grad}
    \begin{split}
    \langle \Delta \theta_t, \nabla l^b_{t} \rangle \approx - a_{xg} \frac{\beta_1^t}{t} + b_{xg} \sqrt {1 - \beta_2 } + \frac{c_{xg}}{t + d_{xg}} \\
    a_{xg}, b_{xg}, c_{xg}, d_{xg} \geq 0, 0 < t < b
    \end{split}
\end{equation}

\begin{equation}
    \label{eq:loss-batch-b}
    \begin{split}
    l^b_t \approx l_0 + \sum_{\tau=0}^{t-1} \langle \Delta \theta_\tau, \nabla l^b_\tau \rangle \\
    0 < t < b
    \end{split}
 \end{equation}

Finally, we address the question of why the Epochal Sawtooth Phenomenon (ESP) effect is weak when data is not shuffled and disappears entirely when samples are drawn with replacement. As noted earlier, when second moment vector $v$ has small components, dividing momentum $\hat{m}$ ($m$ with initialization correction) or gradient by $v$ causes a significant change in direction, often resulting in a new direction that is misaligned with $\nabla l^b$. Our observations suggest that this misalignment interacts with the phenomenon of ``immediate re-exposure to samples'' and ``memorizing the first few batches''. However, we have not yet been able to provide a definitive explanation for how this interaction occurs, or why it disappears when sampling with replacement. One possible explanation is that sampling with replacement improves numerical stability, thereby mitigating the conditions necessary for ESP to manifest.

\section{Experiments}
In this work, we replicate the Epochal Sawtooth Phenomenon using high-dimensional incremental quadratic optimization, demonstrating that this phenomenon is not exclusive to deep learning models but also arises in simpler optimization tasks. This further supports the idea that the observed loss pattern is tied to the inherent characteristics of gradient-based optimization methods. The core part of the code is listed below. We randomly generate 10,000 convex quadratic functions, with the variable $x$ also being of dimension 10,000, mimicking the high capacity of large language models (LLMs). For simplicity, we assume each function acts on a single variable (see code lines 1-6), and each function is randomly assigned an $x$ dimension (see code line 43). The goal is to minimize the sum of all functions (see Eqn \ref{eq:replication-function}). The solution method is the incremental gradient method. The optimizer used is an implementation of \texttt{Adam} (lines 16-35).

\begin{equation}
    \label{eq:replication-function}
    \begin{split}
    F(\Vec{x}) = \sum_{i=1}^{10000} f_i(\Vec{x}) \\
    f_i(\Vec{x}) = c_{i,0}(\Vec{x}[j] - c_{i,1})^2 + c_{i,2} \\
    j \sim \text{DiscreteUniform}\{1, 2, \dots, 10000\} \\
    c_{i,0} \geq 0
    \end{split}
 \end{equation}

\lstset{
  language=Python,
  breaklines=true,
  basicstyle=\footnotesize\ttfamily,  
  keywordstyle=\bfseries\color{blue}, 
  commentstyle=\color{commentgreen}, 
  stringstyle=\color{red},            
  showspaces=false,
  showstringspaces=false,
  frame=single,                       
  numbers=left,                       
  stepnumber=1,                       
  numbersep=5pt,                      
  numberstyle=\tiny\color{gray},      
  tabsize=2                           
}
\begin{lstlisting}
def quadratic_function(x, coeffs, x_dim_idx):
  '''Define a quadratic function acting on one dimension of x'''
  # each func acts on only one variable actually, designated by 'x_dim_idx'
  # coeffs = [a, b, c]
  # return a * (x[x_dim_idx] - b)^2 + c
  return coeffs[0] * (x[x_dim_idx] - coeffs[1])**2 + coeffs[2]

def gradient_quadratic(x, coeffs, x_dim_idx):
  '''Compute gradient of a quadratic function'''
  # coeffs = [a, b, c]
  # return a * (x[x_dim_idx] - b)**2 + c
  grad = np.zeros_like(x)
  grad[x_dim_idx] = 2 * coeffs[0] * (x[x_dim_idx] - coeffs[2])
  return grad

class SimpleAdam:
  '''A simple implementation of Adam optimizer'''
  def __init__(self, lr=0.01, beta_1=0.9, beta_2=0.999, epsilon=1e-8):
    self.lr = lr
    self.beta_1 = beta_1
    self.beta_2 = beta_2
    self.epsilon = epsilon
    self.m = 0  # First moment
    self.v = 0  # Second moment
    self.t = 0  # Time step
        
  def update(self, grad):
    self.t += 1
    self.m = self.beta_1 * self.m + (1 - self.beta_1) * grad
    self.v = self.beta_2 * self.v + (1 - self.beta_2) * (grad ** 2)
    m_hat = self.m / (1 - self.beta_1 ** self.t)
    v_hat = self.v / (1 - self.beta_2 ** self.t)
    # This 't' is not reset. The denominator '(1 - self.beta_1 ** self.t)' is the initialization correction commonly used in Adam optimizer. 
    delta_x = -self.lr * m_hat / (np.sqrt(v_hat) + self.epsilon)
    return delta_x
    
# Settings
num_functions = 10000  # Total number of random quadratic functions
x_dim = 10000  # Dimensionality of input vector x
# Coefficients for random functions; all functions are convex
coeffs_list = np.random.uniform(0.5, 1.0, size=(num_functions, 3))  
coeffs_list[:, 1] = np.random.uniform(-1, 1, size=num_functions) # Minima near zero
x_dim_indices = np.random.randint(0, x_dim, size=num_functions) # Assign each function to act on a random dimension of x
x_init = 3.0 * np.ones(x_dim)  # Initial point
optimizer = SimpleAdam(lr=0.06, beta_1=0.9, beta_2=0.999)
\end{lstlisting}

\begin{figure}
\centering
\includegraphics[width=1.0\linewidth]{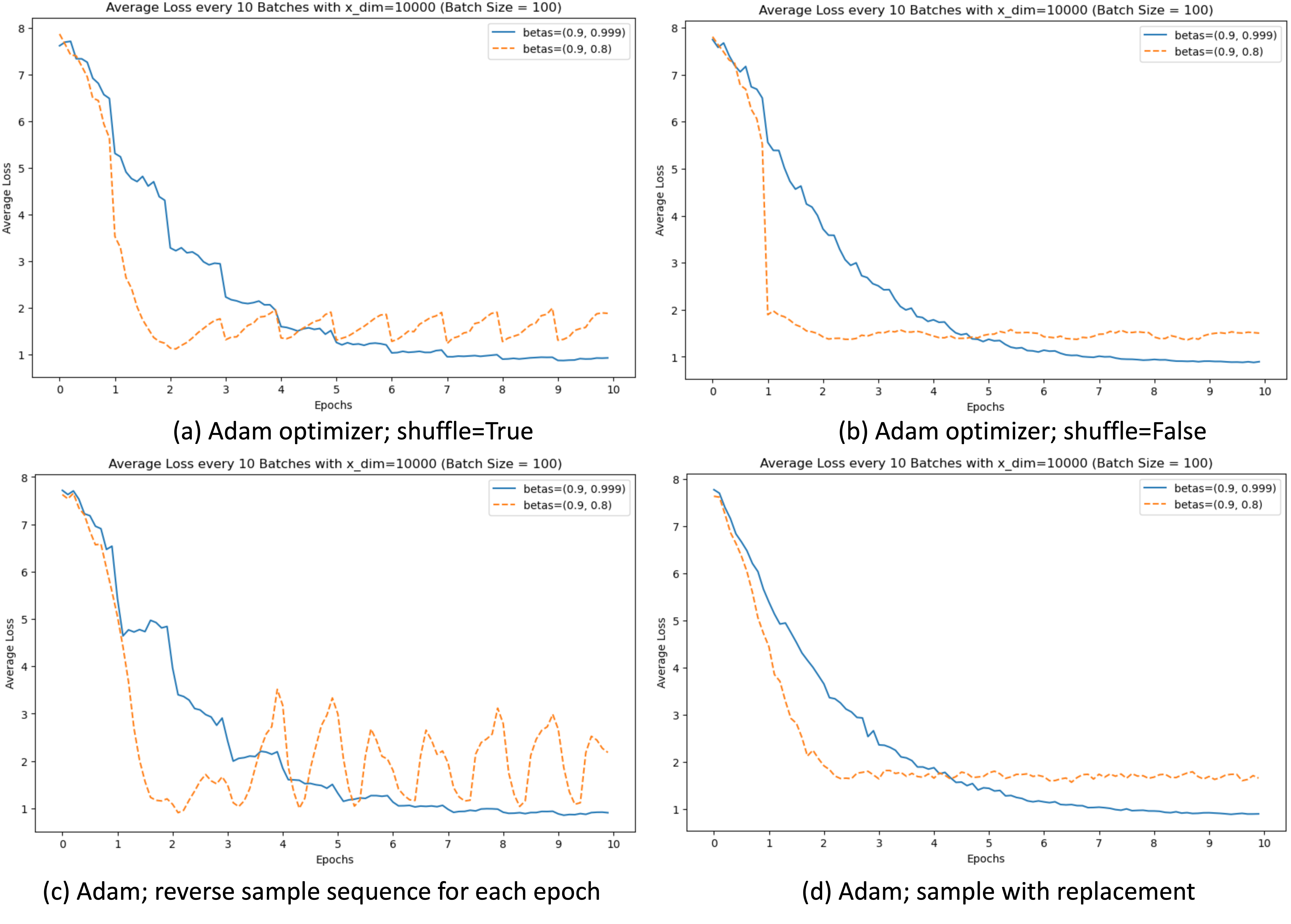}
\caption{The effects of data shuffling on incremental quadratic optimization. (a) Incremental quadratic optimization with \texttt{shuffle=True}. ESP is replicated when we shuffle data. Smaller $\beta_2$ exacerbates ESP. Similar to Figure \ref{fig:beta-2} (b) When \texttt{Shuffle=False}, ESP is not observed. (c) Reverse the sample sequence for each epoch. This significantly amplifies the ESP, aligning with our earlier analysis illustrated in Figure \ref{fig:sequence-momentum-illust}. (d) Sample with replacement.  ESP is not observed.
}
\label{fig:replication-benchmark}
\end{figure}

The results are presented in Figure \ref{fig:replication-benchmark}. The Epochal Sawtooth Phenomenon (ESP) is successfully replicated when the data is shuffled (Figure \ref{fig:replication-benchmark} (a)). It is absent when the data is not shuffled (Figure \ref{fig:replication-benchmark} (b)) or when sampling is performed with replacement (Figure \ref{fig:replication-benchmark} (d)). Furthermore, we experimented with reversing the sample sequence for each epoch using an “AB, BA, AB…” sampling strategy, as described in the previous section. This approach significantly amplifies the ESP, as shown in Figure \ref{fig:replication-benchmark} (c), which aligns with our earlier analysis presented in Figure \ref{fig:sequence-momentum-illust}. Additionally, the randomness in the training loss curve oscillations are noticeably more pronounced under this strategy.

ESP is more pronounced when $\beta_2$ is small and transitions to a staircase pattern as $\beta_2$ approaches 1 (see Figure \ref{fig:replication-benchmark} (a)). We experimented with different batch sizes and found that a larger batch size can mitigate ESP, similar to our previous findings. We also replaced Adam with RMSProp, another widely used adaptive optimizer. ESP still exists, but it is much more subtle. At epoch 10, the training loss increased by about 10\% at the end of the epoch. Additionally, we found that no-shuffling mitigates ESP in a manner similar to Adam (See additional figures on Github).

\begin{figure}
    \centering
    \includegraphics[width=1.0\linewidth]{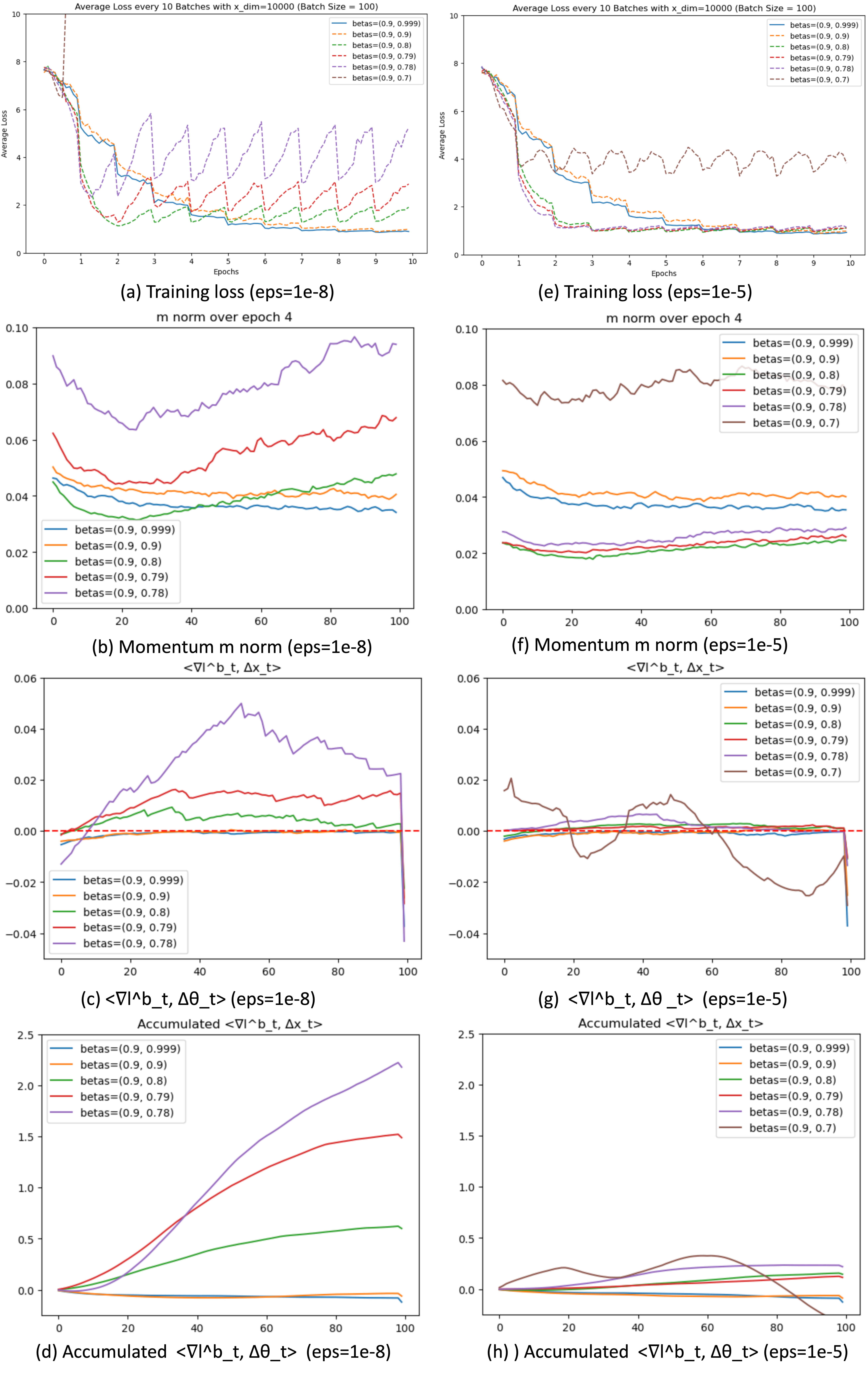}
    \caption{The effects of varying $\beta_2$ and $\epsilon$ (a)-(d) Training loss, momentum norm, $\langle \nabla l^b_t, \Delta \theta_t \rangle$, and accumulated $\langle \nabla l^b_t, \Delta \theta_t \rangle$ under $\epsilon=1 \times 10^{-8}$; The training becomes unstable when $\beta_2 < 0.8$ and diverges at $\beta_2 = 0.7$; (e)-(h) Training loss, momentum norm, $\langle \nabla l^b_t, \Delta \theta_t \rangle$, and accumulated $\langle \nabla l^b_t, \Delta \theta_t \rangle$ under $\epsilon=1 \times 10^{-5}$. Large $\epsilon$ in Adam can stabilize training but does not necessarily restore optimality.} 
    \label{fig:vary_beta2_and_eps}
\end{figure}

We also observed that once $\beta_2$ becomes less than 0.8, the training process quickly becomes unstable. At $\beta_2 = 0.7$, the algorithm diverges, as shown in Figure \ref{fig:vary_beta2_and_eps} (a). This observation aligns with the recommended settings for $\beta_2$ in LLM training: values of $\beta_2$ below 0.8 are prone to instability and are considered too small. Additionally, ESP becomes less pronounced as $\beta_1$ decreases in this example. This feature differs from the BERT example (See additional figures on Github) 

We also explored how the value of $\epsilon$ in $\texttt{Adam}$ affects training. The default value of $\epsilon$ is $1 \times 10^{-8}$, but when we changed it to $1 \times 10^{-5}$, we observed that training became more stable, as shown in Figure \ref{fig:vary_beta2_and_eps} (b). The cosine similarity sign-flipping effect observed earlier was mitigated. The algorithm converged even at $\beta_2 = 0.7$, though it converged to an inferior value. This suggests that a larger $\epsilon$ in Adam can stabilize training, but it does not necessarily restore optimality.

We also compare the losses $l^t_t$ and $l^b_t$ for a fixed $b = 100$ over epoch 4 (See additional figures on Github). We observe that they exhibit the same pattern, which verifies Eqn \ref{eq:loss-batch-b}. Finally, we note that one limitation of this replication is that the concavity of the loss curve for small values of $\beta_2$ was not reproduced. This may be a feature unique to deep neural networks.

\section{Conclusion}
In this paper, we identify and analyze a recurring loss pattern, which we term the Epochal Sawtooth Phenomenon (ESP), frequently observed during training with adaptive gradient-based optimizers, particularly Adam. This phenomenon is characterized by a sharp decrease in loss at the start of each epoch, followed by a gradual increase. Through both theoretical and empirical investigation, we demonstrate that ESP emerges due to several interacting factors, including data reshuffling, the configuration of Adam’s parameters, batch size, and model capacity. Specifically, the ``immediate re-exposure to samples'' effect, induced by data shuffling, causes the model to learn more (or memorize) at the beginning of each epoch. Smaller values of $\beta_2$ exacerbate ESP, though they may also serve as a form of regularization. While ESP does not necessarily indicate overfitting, higher model capacity tends to amplify the phenomenon.

We present a quantitative analysis of the loss dynamics throughout an epoch, deriving the training loss as a function of optimizer parameters and training steps. Our analysis shows how gradient-based updates are modulated by Adam’s adaptive learning rate mechanism, with $\beta_2$ playing a pivotal role in shaping the slope and curvature of the loss curve. Specifically, we fit a formula for the training loss, demonstrating how the combination of optimizer parameters and gradient statistics dictates the evolution of the loss over time.

Furthermore, we replicate ESP using a quadratic minimization task, reinforcing its generality across different optimization settings. By solving a sequence of quadratic problems, we show that this phenomenon persists even in simplified environments, suggesting that ESP is not unique to neural networks but rather a broader characteristic of gradient-based optimization.

One potential direction for future work is to investigate ESP in simpler analytical settings, such as linear or logistic regression. This could facilitate more rigorous theoretical derivations and complement the empirical observations presented in this work.
It's also helpful to refine the replication model and formulas to account for the concavity observed when $\beta_2$ is small. Additionally, further exploration of how ESP impacts generalization and model robustness could lead to more efficient training strategies for large-scale deep learning systems.

\section{Acknowledgment}
We would like to express our sincere gratitude to Editor Michel Verleysen and the anonymous reviewers for their valuable insights and constructive feedback, which significantly contributed to the improvement of this study. The work described in this paper was partly supported by research grant from Shanghai Baiyulan Talent Program Pujiang Project under grant number 24PJD115.



\end{document}